\documentclass{article}

\usepackage{microtype}
\usepackage{graphicx}
\usepackage{subfigure}
\usepackage{booktabs} 

\usepackage{hyperref}

\usepackage[accepted]{icml2025}

\usepackage{amsmath}
\usepackage{amssymb}
\usepackage{mathtools}
\usepackage{amsthm}

\usepackage[capitalize,noabbrev]{cleveref}

\theoremstyle{plain}

\theoremstyle{definition}

\theoremstyle{remark}

\usepackage[textsize=tiny]{todonotes}

\icmltitlerunning{Foreground-aware Virtual Staining for Accurate 3D Cell Morphological Profiling}

\begin{document}

\twocolumn[
\icmltitle{Foreground-aware Virtual Staining for Accurate 3D Cell Morphological Profiling}




\begin{icmlauthorlist}
\icmlauthor{Alexandr A.~Kalinin}{broad,dcmb}
\icmlauthor{Paula Llanos}{broad}
\icmlauthor{Theresa~Maria Sommer}{imba}
\icmlauthor{Giovanni Sestini}{imba}
\icmlauthor{Xinhai Hou}{dcmb}
\icmlauthor{Jonathan Z.~Sexton}{dim,dmc}
\icmlauthor{Xiang Wan}{sribd}
\icmlauthor{Ivo Dinov}{dcmb,socr,nursing}
\icmlauthor{Brian D.~Athey}{dcmb}
\icmlauthor{Nicolas Rivron}{imba}
\icmlauthor{Anne E.~Carpenter}{broad}
\icmlauthor{Beth Cimini}{broad}
\icmlauthor{Shantanu Singh}{broad}
\icmlauthor{Matthew J.~O’Meara}{dcmb,dmc}
\end{icmlauthorlist}

\icmlaffiliation{broad}{Imaging Platform, Broad Institute of MIT and Harvard, Cambridge, MA 02142, USA}
\icmlaffiliation{dcmb}{Gilbert S.~Omenn Department of Computational Medicine and Bioinformatics, University of Michigan Medical School, Ann Arbor, MI 48109, USA}
\icmlaffiliation{imba}{Institute of Molecular Biotechnology of the Austrian Academy of Sciences (IMBA), Vienna BioCenter (VBC), 1030 Vienna, Austria}
\icmlaffiliation{dim}{Department of Internal Medicine – Gastroenterology, University of Michigan Medical School, Ann Arbor, MI 48109, USA}
\icmlaffiliation{dmc}{Department of Medicinal Chemistry, College of Pharmacy, University of Michigan, Ann Arbor, MI 48109, USA}
\icmlaffiliation{sribd}{Shenzhen Research Institute of Big Data, Chinese University of Hong Kong–Shenzhen, Shenzhen 518172, Guangdong, China}
\icmlaffiliation{socr}{Statistics Online Computational Resource (SOCR), University of Michigan, Ann Arbor, MI 48109, USA}
\icmlaffiliation{nursing}{Health Behavior and Biological Sciences, University of Michigan School of Nursing, Ann Arbor, MI 48109, USA}

\icmlcorrespondingauthor{Alexandr A. Kalinin}{akalinin@broadinstitute.org}
\icmlcorrespondingauthor{Matthew J. O’Meara}{momeara@umich.edu}

\icmlkeywords{Machine Learning, ICML}

\vskip 0.3in
]



\printAffiliationsAndNotice{}  

\begin{abstract}
Microscopy enables direct observation of cellular morphology in 3D, with transmitted-light methods offering low-cost, minimally invasive imaging and fluorescence microscopy providing specificity and contrast. Virtual staining combines these strengths by using machine learning to predict fluorescence images from label-free inputs. However, training of existing methods typically relies on loss functions that treat all pixels equally, thus reproducing background noise and artifacts instead of focusing on biologically meaningful signals. We introduce Spotlight, a simple yet powerful virtual staining approach that guides the model to focus on relevant cellular structures. Spotlight uses histogram-based foreground estimation to mask pixel-wise loss and to calculate a Dice loss on soft-thresholded predictions for shape-aware learning. Applied to a 3D benchmark dataset, Spotlight improves morphological representation while preserving pixel-level accuracy, resulting in virtual stains better suited for downstream tasks such as segmentation and profiling.
\end{abstract}

\section{Introduction}
\label{introduction}

Microscopy is an essential tool for studying cellular states, enabling direct observation of cell morphology and dynamics with high spatial and temporal resolution. Transmitted-light microscopy has low cost, straightforward implementation, and minimal invasiveness, which make it ideal for long-term and live-cell imaging. However, it is limited by the inherently low specificity and contrast. Fluorescence microscopy dramatically improves contrast and enables multiplexed imaging of several biological structures simultaneously by utilizing fluorescent probes that selectively label cellular targets. However, fluorescence microscopy requires more complex and costly instrumentation and sample preparation, frequently induces phototoxicity and photobleaching, and usually requires samples to be fixed, precluding studies of living samples over time. Moreover, the limited availability of fluorescence spectra restricts its utility for multiplexing. These limitations motivated the development of computational methods collectively termed "virtual staining" or "in silico labeling" \cite{Ounkomol2018-wa,Christiansen2018-xn}. Virtual staining leverages machine learning (ML) algorithms to predict fluorescent signals from transmitted-light microscopy images, aiming to integrate the simplicity, minimal invasiveness, and affordability of transmitted-light methods with the specificity and contrast of fluorescence imaging. Thus, virtual staining approaches offer a promising strategy to enhance throughput, minimize experimental variability, and facilitate longer-term, minimally invasive observation of live cells, significantly expanding the applicability and value of microscopy-based cellular studies.

\begin{figure*}[t]
\vskip 0.2in
\begin{center}
\centerline{\includegraphics[width=\textwidth]{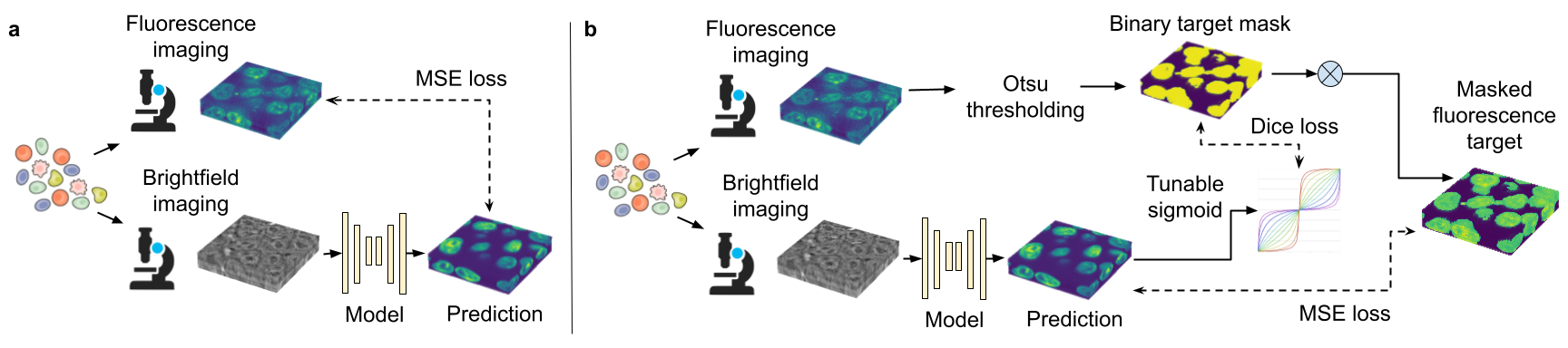}}
\caption{A schematic overview of Spotlight. a: Typical virtual staining models are trained using a pixel-wise loss such as mean squared error (MSE) on the whole image. b: Spotlight uses foreground estimation obtained by histogram thresholding to restrict pixel-wise loss to foreground areas and also employs soft-thresholding of the prediction to compute segmentation loss.}
\label{fig1}
\end{center}
\vskip -0.2in
\end{figure*}

Virtual staining techniques originated from image-to-image translation research in computer vision, where models map paired image domains such as sketches to photographs, or day scenes to night scenes \cite{Isola2017-fy}. Adopting this idea for virtual staining, Ounkomol et al.~\yrcite{Ounkomol2018-wa} and Christiansen et al.~\yrcite{Christiansen2018-xn} introduced U-net–like models \cite{Ronneberger2015-ey} trained with pixel-wise losses for predicting fluorescence signals for multiple subcellular structures in 3D from brightfield or phase-contrast stacks, establishing the feasibility of the overall approach. Subsequent works have experimented with input modalities and custom imaging setups \cite{Cheng2021-iu,Guo2020-xc} and enhanced architectures \cite{Wang2021-xr,Zhou2023-su}, but the core training scheme remained supervised regression against real fluorescence images. The training objectives in most of these models are dominated by pixel-wise loss functions such as mean squared error (MSE) or mean absolute error (MAE), sometimes supplemented with perceptual or structural similarity terms \cite{liu2025robust}. Adversarial losses have been introduced to improve realism, yet they are coupled with per-pixel MSE/MAE terms to stabilize training \cite{Rivenson2019-ks,Wang2023-hr}.

These approaches overlook a fundamental difference between biological microscopy images and natural images: translating transmitted-light microscopy into fluorescence images is not merely stylistic but must accurately capture underlying biological structures. Unlike natural images, where even background pixels typically convey semantic information such as representing the sky, trees, or buildings, microscopy images often have biologically irrelevant backgrounds. Because most existing virtual staining models are trained with pixel-wise losses (\cref{fig1}a), they treat all pixels as equally important and learn to reproduce noise and artifacts from biologically irrelevant background regions of original fluorescence images, making downstream analysis challenging. This limitation becomes even more critical in 3D imaging, where axial aberrations and elongation distort cellular structures. As a result, predicted fluorescence images often inherit the segmentation difficulties of the original fluorescence data, complicating the segmentation of individual cells, which is a common downstream task.

Here, we introduce a new method, termed Spotlight, to address these challenges by explicitly guiding virtual staining models to focus on predicting cellular structures in the foreground (FG), while ignoring background (BG). Spotlight incorporates two key components (\cref{fig1}b). First, we apply a simple histogram thresholding procedure to approximate the location of FG structures, using this to mask traditional pixel-wise loss functions and restrict supervision to informative regions of the image. Second, we encourage accurate modeling of cell shape by soft-thresholding the predicted fluorescence output and computing a segmentation loss with respect to the FG mask. The combined training objective balances precise FG intensity prediction with geometric fidelity, enabling the model to generate morphologically accurate virtual stains, thereby facilitating downstream tasks such as segmentation and profiling. We demonstrate Spotlight on a benchmark dataset and show that it improves 3D morphological representation compared to standard training approaches.

\section{Spotlight}
\label{spotlight}

Training virtual staining models with pixel-wise loss functions is especially limiting in 3D microscopy, where the background dominates the voxel count, while containing noise and optical aberration artifacts without real biological signal. Our main insight is that even a weak approximation of FG is sufficient to prioritize training on meaningful content and suppress irrelevant noise from BG regions.  

\begin{figure*}[t]
\vskip 0.2in
\begin{center}
\centerline{\includegraphics[width=\textwidth]{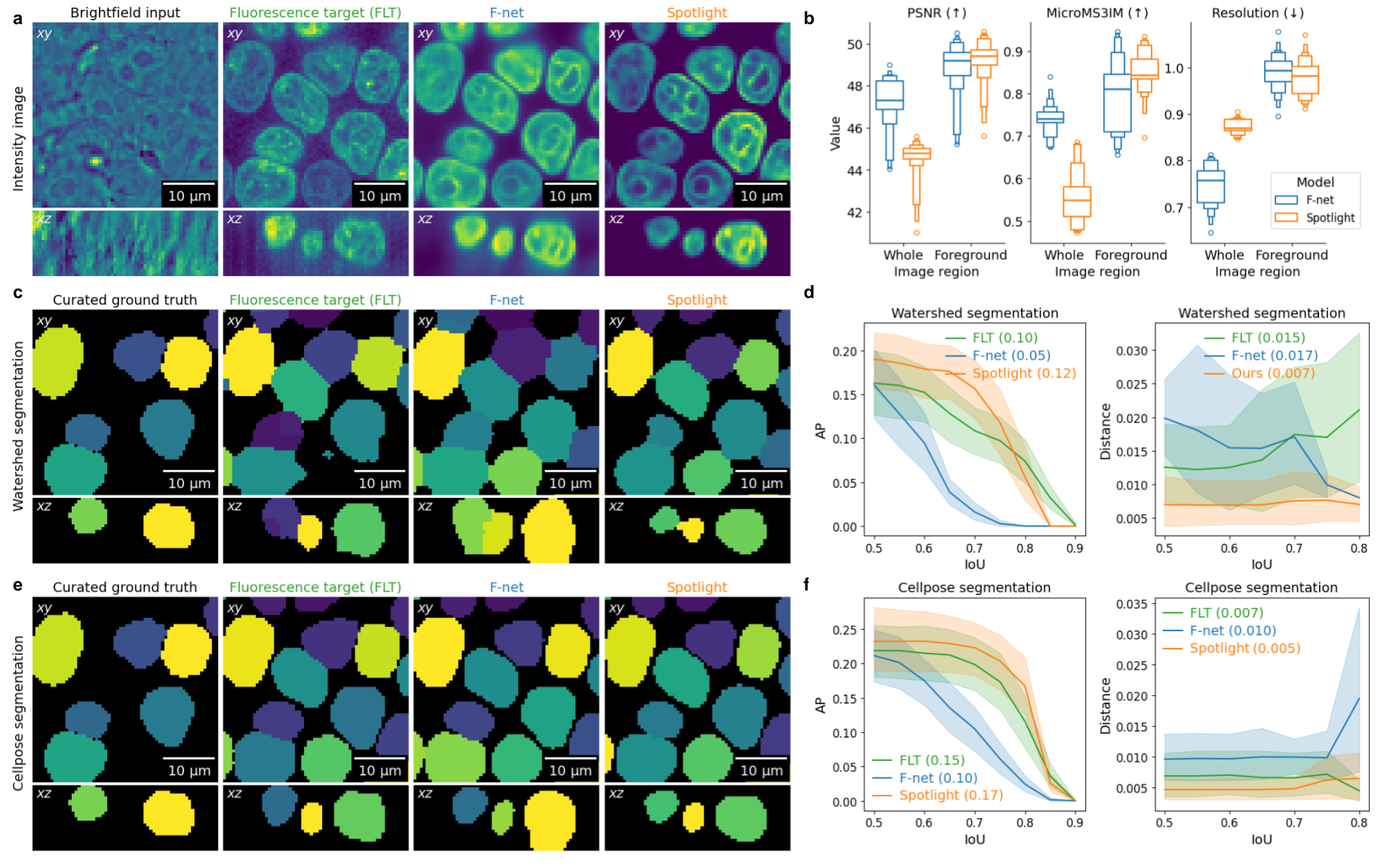}}
\caption{Performance evaluation on the Allen Cell Label-free Imaging dataset. a: Visualizations of F-net and Spotlight predictions demonstrate differences in BG prediction. b: Image quality metrics indicate that Spotlight preserves the same level of detail in FG intensities. c,e: Visualizations of wathershed and Cellpose segmentation results show artifacts when segmenting F-net predictions. d,f: Segmentation and feature-level metrics indicate that less noisy predictions by Spotlight produce more accurate morphologies.}
\label{fig2}
\end{center}
\vskip -0.2in
\end{figure*}

Two components of Spotlight incorporates use the FG approximation to guide the training process (\cref{fig1}b). The first component applies FG mask to a traditional pixel-wise loss function (e.g., MSE), thereby restricting supervision to pixels likely to correspond to cellular structures. The second soft-thresholds the model’s predicted output and computes Dice loss \cite{Milletari2016-ap} against the binary FG mask, encouraging preservation of shape and structure in the predicted fluorescence. The masked pixel loss enables the model to ignore uninformative BG regions, while the Dice loss promotes spatial coherence and reduces noise and artifacts in predictions. Together, these losses create a training signal that emphasizes morphological fidelity without compromising pixel-level accuracy.

\textbf{Foreground estimation:} To restrict supervision to biologically informative regions, Spotlight uses a FG mask derived from the target fluorescence image using Otsu’s method~\cite{Otsu1979-eg}.

Let $Y$ denote the ground truth fluorescence image, $\hat{Y}$ the predicted image, and $T$ the Otsu threshold computed from the histogram of $Y$ as the intensity value that maximizes inter-class variance. We then define the binary foreground mask $M \in {0,1}^{H \times W \times D}$ by thresholding $Y$ at $T$, such that $M_{i,j,k} = 1$ if $Y_{i,j,k} \geq T$, and $0$ otherwise. Using this mask, we compute the masked mean squared error (MMSE) loss:
\[
\mathcal{L}_{\text{MMSE}} = \frac{1}{\sum M} \sum_{i,j,k} M_{i,j,k} \left(Y_{i,j,k} - \hat{Y}_{i,j,k}\right)^2
\]

\begin{figure*}[t]
\vskip 0.2in
\begin{center}
\centerline{\includegraphics[width=\textwidth]{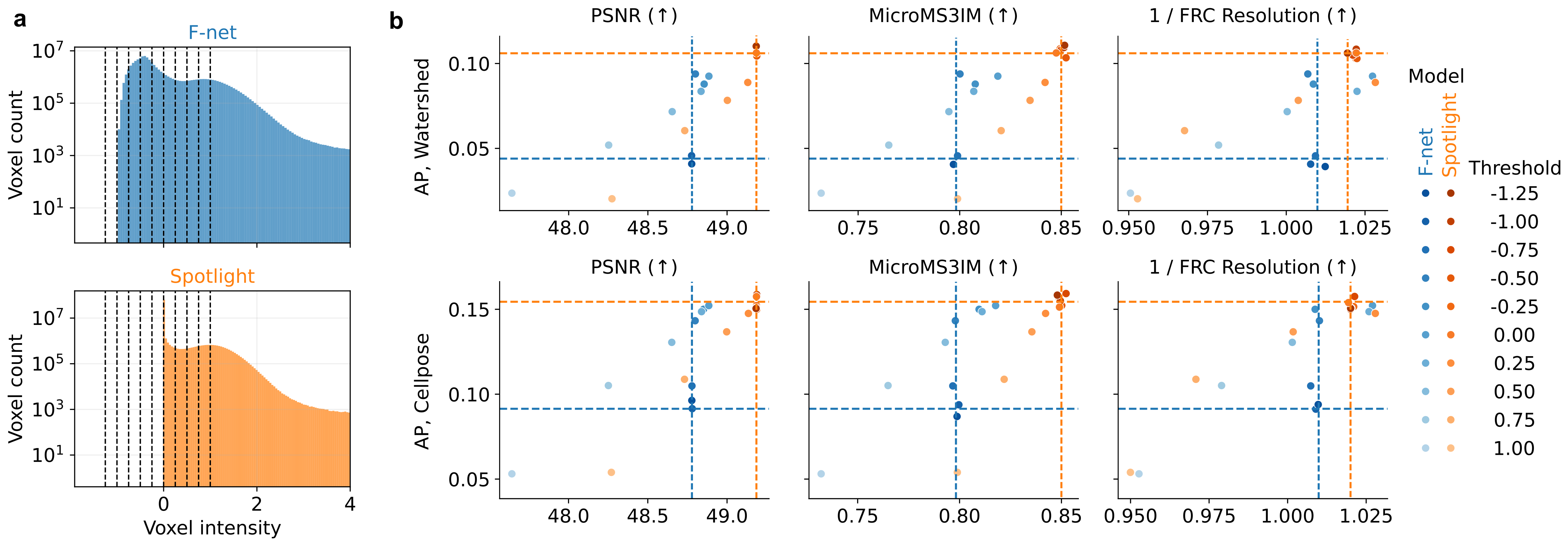}}
\caption{Performance comparison on thresholded predictions. a: Intensity histograms for F-net and Spotlight test predictions. (black dashed lines: tested post-processing thresholds). b: Segmentation and image-quality metrics calculated on thresholded F-net and Spotlight predictions. Colored dashed lines correspond to evaluation of F-net and Spotlight predictions without any thresholding.}
\label{fig3}
\end{center}
\vskip -0.2in
\end{figure*}

\textbf{Morphology-preserving component:} To encourage morphological fidelity, we apply a differentiable soft-thresholding function to the prediction. We use a normalized tunable sigmoid~\cite{dhemery_sigmoid} defined as:
\[
\sigma_k(x) = \frac{x - kx}{k - 2k|x| + 1}
\]
where $k$ controls sharpness and nonlinearity.

Applying this to $\hat{Y}$, we compute the Dice loss against the binary foreground mask:
\[
\mathcal{L}_{\text{Dice}} = 1 - \frac{2 \sum \sigma_k(\hat{Y}) \cdot M}{\sum \sigma_k(\hat{Y}) + \sum M + \epsilon}
\]

The total loss used for training is a weighted combination:
\[
\mathcal{L}_{\text{Spotlight}} = \lambda \cdot \mathcal{L}_{\text{MMSE}} + (1 - \lambda) \cdot \mathcal{L}_{\text{Dice}}
\]
This loss formulation emphasizes accurate reconstruction of informative regions and structurally coherent fluorescence predictions.

\section{Experiments}
\label{experiments}

\subsection{Experimental setup}
\label{experimental_setup}

We used the “DNA” subset (Hoechst-lebeled nuclei) of the Allen Institute for Cell Science Label-free Determination dataset~\cite{Ounkomol2018-wa}.
As a baseline, we re-trained the commonly used 3D F-net model proposed by Ounkomol et al.~\yrcite{Ounkomol2018-wa}.
We evaluated predictions at the pixel, segmentation, and feature levels.
At the pixel level, we compute prediction image quality metrics both on the whole image and only the foreground, including peak signal-to-noise ratio (PSNR), microscopy-specific 3D structural similarity (MicroMS3IM)~\cite{Ashesh2025-ll}, and the Fourier ring correlation-based (FRC) image resolution estimate~\cite{Koho2019-cx}.
We tested two separate segmentation approaches: Otsu’s thresholding~\cite{Otsu1979-eg} followed by the watershed algorithm~\cite{Soille1990-pj} and Cellpose with the pre-trained “nuclei” model~\cite{Stringer2021-qk}, and reported average precision (AP) calculated at different intersection over union (IoU) levels against the manually curated ground-truth segmentation masks provided with the dataset~\cite{Ounkomol2018-wa}.
At the feature level, we report cosine distance between 3D measurement profiles extracted from the target fluorescence images and profiles extracted from the model predictions.

\subsection{Results}
\label{results}
Our results show that incorporating a foreground-aware loss substantially improves virtual staining of nuclei compared to the baseline F-net by reducing artifacts such as axial elongation (\cref{fig2}). Trained with MSE loss, F-net predictions are smoothed compared to the original fluorescence images (\cref{fig2}a). While this suppressed noise, cell edges remain blurred and the “halo” of axial elongation is clearly present. In contrast, Spotlight preserves nuclear boundaries and produces predictions free of these artifacts (\cref{fig2}a). Quantitatively, image‐quality metrics worsen when computed over the entire volume but are consistent when restricted to the FG mask (\cref{fig2}b). This improved FG/BG separation translates directly into more accurate segmentations—using both watershed (\cref{fig2}c) and Cellpose(\cref{fig2}e)—and yields morphological measurements that align more closely with the ground truth (\cref{fig2}d,f).

We also compared the voxel‐intensity distributions of F-net and Spotlight predictions (\cref{fig3}a). Whereas F-net produces a clear bimodal BG/FG histogram, Spotlight sharply collapses background variability toward zero, creating a tighter separation between background and foreground. To test how this affects segmentation, we thresholded each prediction volume at a range of intensity values spanning both BG/FG modes, and then ran two segmentation pipelines (watershed and Cellpose). For F-net, filtering out BG intensities improved both segmentation and image-quality scores—bringing them nearly up to Spotlight’s levels—whereas Spotlight’s metrics stayed consistent (\cref{fig3}b). We show how learning from informative foreground regions in virtual staining acts allows to avoid complicated post-processing, to simplify segmentation and to improve its utility for 3D nuclear morphometry.

\section{Conclusion}
Less invasive, cost-effective computational imaging will be valuable for a wide range of applications in biomedicine, such as live-cell assays. Methods that do not compromise on morphological detail and preserve cellular geometry and structure can be helpful for downstream tasks such as segmentation, phenotyping, and profiling. As ML methods continue to shape biological imaging, it is essential to consider the statistical, structural, and physical properties of the data—not just to improve prediction accuracy, but to ensure that models yield biologically relevant results.

\section*{Acknowledgements}

This work was pertially supported by Chinese Key-Area Research and Development Program of Guangdong Province (2020B0101350001).
This work was partially supported by the Human Frontier Science Program (RGY0081/2019 to S.S.) and a grant from the National Institutes of Health NIGMS (R35 GM122547 to A.E.C.).
Xin Rong of the University of Michigan donated NVIDIA TITAN X GPU used for this research, and the NVIDIA Corporation donated the TITAN Xp GPU used to execute the computationally intensive protocol.

\section*{Impact Statement}
This work aims to reduce the cost, phototoxicity, and reagent use of fluorescence microscopy by generating virtual stains from label-free images. This can accelerate high-content screening and make advanced imaging more accessible in resource-limited labs. Because downstream biological conclusions may rely on the fidelity of virtual stains, inaccurate predictions could mislead research or clinical workflows if used uncritically. We therefore advocate reporting rigorous benchmarks so that users can assess reliability in their own settings before deployment.


\bibliography{example_paper}
\bibliographystyle{icml2025}

\newpage
\appendix

\section{Appendix}

\subsection{Dataset details}
\label{experimental_setup_data}

We used the “DNA” subset (Hoechst-lebeled nuclei) of the Allen Institute for Cell Science Label-free Determination dataset~\cite{Ounkomol2018-wa}. This dataset includes manually curated ground truth segmentation masks, which we use to evaluate the segmentation quality of nuclei. Masks that did not pass quality control were removed, which led to some cells missing the ground truth segmentation masks (see \cref{fig2}c, "Curated ground truth").

\subsection{Baseline details}
\label{experimental_setup_baseline}

\textbf{F-net:} F-net~\cite{Ounkomol2018-wa} is a fully convolutional U-Net variant that consists of layers that perform one of three convolution types, followed by a batch normalization and rectified linear unit (ReLU) operation. The convolutions are either 3-pixel convolutions with a stride of 1 pixel on zero-padded input (such that the input and output of that layer are the same spatial area), 2-pixel convolutions with a stride of 2 pixels (to halve the spatial area of the output), or 2-pixel transposed convolutions with a stride of 2 pixels (to double the spatial area of the output). There are no normalization or ReLU operations on the last layer of the network.

\subsection{Training and evaluation}
\textbf{Training protocol:} We followed the original data split, preprocessing, and training protocols from Ounkomol et al.~\yrcite{Ounkomol2018-wa}. Specifically, we resized all z-stacks using cubic interpolation to achieve isotropic voxel dimensions of 0.29×0.29×0.29µm³. We trained models on batches of 3D patches (32×64×64 px³, ZYX), which were randomly subsampled uniformly both across all training images and spatially within an image. All models were trained using the Adam optimizer~\cite{Kingma2015-ni} with a learning rate of 0.001, a batch size of 24, and with beta values of 0.5 and 0.999 for 50,000 mini-batch iterations. All models were implemented in Python using the PyTorch package~\cite{Paszke2019-jb}.

\textbf{Evaluation protocol:}
Evaluation followed Ounkomol et al.~\yrcite{Ounkomol2018-wa}, including cropping the input image such that its size in any dimension is a multiple of 16.

When calculating pixel-level image quality metrics, predicted intensities were rescaled into the original fluorescence image intensity range using maximum and minimum values of the combined training set. The FRC resolution was calculated using the miplib package~\cite{Koho2019-cx} with image zero-padding and parameter bin\_delta=5.

Segmentation was performed on images downscaled by half in all dimensions. Hyperparameters for both segmentation pipelines were chosen to optimize segmentation results on the real fluorescence images. Watershed segmentation roughly followed the CellProfiler's \textit{3d\_monolayer} tutorial~\cite{stirling2021cellprofiler}, with the additional step of contrast-limiting adaptive histogram equalization~\cite{pizer1987adaptive}. Segmentation post-processing included removing nuclei at image edges, filling holes, and filtering out nuclei by size at the empirically chosen cutoffs.

Morphological feature extraction was performed using a combination of \texttt{regionprops} method from scikit-image~\cite{van2014scikit} and mesh properties~\cite{kalinin2021valproic}.

Implementation of the evaluation protocol relied on popular Python libraries, including numpy~\cite{harris2020array}, scipy~\cite{virtanen2020scipy}, pandas~\cite{mckinney-proc-scipy-2010}, and RAPIDS cuCIM~\cite{cucim}. Figures were made using matplotlib~\cite{Hunter_2007} and seaborn~\cite{Waskom2021}.

\subsection{Spotlight details}
Spotlight uses the same model architecture as F-net. The only differences are introduced in data preprocessing, patch selection, and the loss function.

\textbf{Preprocessing:} While F-net training standardizes 3D patches using mean and standard deviation, we replaced mean with the threshold value calculated using Otsu's method to simplify soft-thresholding of predictions in the calculation of the Dice loss.

\textbf{Patch selection:} In order to facilitate foreground-aware training, we also implemented patch selection during training such that only 3D patches containing at least 0.1\% of foreground voxels were used for training.

\textbf{Loss function:} The overall loss function design is described in \cref{fig1} and \cref{spotlight}. We empirically set sharpness parameter $k$ in the normalized tunable sigmoid function equal to -0.95 and weight $\lambda$ equal to 0.5.


\end{document}